\documentclass{article}

\usepackage{amsmath}
\usepackage{amssymb}
\usepackage{amsthm}
\usepackage{todonotes}
\usepackage{graphicx}
\usepackage{hyperref}
\usepackage{natbib}
\usepackage{algorithm}
\usepackage{algorithmic}

\title{Addiplication}
\author{Sebastian Urban, Patrick van der Smagt}

\newcommand{\setN}{\mathbb{N}}
\newcommand{\setZ}{\mathbb{Z}}
\newcommand{\setR}{\mathbb{R}}
\newcommand{\setC}{\mathbb{C}}
\newcommand{\e}{\mathrm{e}}
\renewcommand{\vec}[1]{\boldsymbol{#1}}
\newcommand{\rulesep}{\unskip\ \vrule\ }
\renewcommand{\Re}{\mathrm{Re}\,}
\renewcommand{\Im}{\mathrm{Im}\,}

\usepackage[accepted]{icml2013} 

\icmltitlerunning{A Smooth and Differentiable Transition Between Additive and Multiplicative Interactions}

\begin{document}


\twocolumn[
\icmltitle{A Neural Transfer Function for a Smooth and Differentiable Transition Between Additive and Multiplicative Interactions}

\icmlauthor{Sebastian Urban}{surban@tum.de}
\icmladdress{Institut f\"ur Informatik VI, Technische Universit\"at M\"unchen, Boltzmannstr. 3, 85748 Garching, Germany}
\icmlauthor{Patrick van der Smagt}{smagt@brml.org}
\icmladdress{fortiss --- An-Institut der Technischen Universit\"at M\"unchen, Guerickestr. 25, 80805 M\"unchen, Germany}

\icmlkeywords{neural nets, machine learning, Abel, Schroeder, interpolation, transfer function, multiplicative, functional iteration, spatial translation}

\vskip 0.3in
]

\begin{abstract}
Existing approaches to combine both additive and multiplicative neural units either use a fixed assignment of operations or require discrete optimization to determine what function a neuron should perform.
This leads either to an inefficient distribution of computational resources or an extensive increase in the computational complexity of the training procedure.

We present a novel, parameterizable transfer function based on the mathematical concept of non-integer functional iteration that allows the operation each neuron performs to be smoothly and, most importantly, differentiablely adjusted between addition and multiplication. 
This allows the decision between addition and multiplication to be integrated into the standard backpropagation training procedure.
\end{abstract}

\section{Introduction}
 
In commonplace artificial neural networks (ANNs) the value of a neuron is given by a weighted sum of its inputs propagated through a non-linear transfer function.
For illustration let us consider a simple neural network with multidimensional input and multivariate output.
The input layer should be called $\vec{x}$ and the outputs $\vec{y}$.
Then the value of neuron $y_i$ is
\begin{equation}
y_i = \sigma\!\left( \sum_j W_{ij} x_j  \right) \,. 
\label{eq:additivenn}
\end{equation}
The typical choice for the transfer function $\sigma(t)$ is the sigmoid function $\sigma(t) = 1/(1+\e^{-t})$ or an approximation thereof.
Matrix multiplication is used to jointly compute the values of all neurons in one layer more efficiently; we have 
\begin{equation}
\vec{y} = \vec{\sigma}( W \vec{x} ) 
\label{eq:nnaddmat}
\end{equation}
where the transfer function is applied element-wise, $\vec{\sigma}_i(\vec{t}) = \sigma(t_i)$.
In the context of this paper we will call such networks additive ANNs.

\cite{hornik1989multilayer} showed that additive ANNs with at least one hidden layer and a sigmoidal transfer function are able to approximate any function arbitrarily well given a sufficient number of hidden units.
Even though an additive ANN is an universal function approximator, there is no guarantee that it can approximate a function \emph{efficiently}.
If the architecture is not a good match for a particular problem, a very large number of neurons is required to obtain acceptable results.

\cite{Durbin1989} proposed an alternative neural unit in which the weighted summation is replaced by a product, where each input is raised to a power determined by its corresponding weight.
The value of such a \emph{product unit} is given by
\begin{equation}
y_i = \sigma\!\left( \prod_j x_j^{W_{ij}}  \right) \,. 
\label{eq:multiplicativenn}
\end{equation}
Using laws of the exponential function this can be written as
$y_i = \sigma [ \exp( \sum_j W_{ij} \log x_j  ) ]$
and thus the values of a layer can also be computed efficiently using matrix multiplication, i.e.
\begin{equation}
\vec{y} = \vec{\sigma}(\exp( W \log \vec{x} ) )  
\label{eq:nnmultmat}
\end{equation}
where $\exp$ and $\log$ are taken element-wise.
Since in general the incoming values $\vec{x}$ can be negative, the complex exponential and logarithm are used.
Often no non-linearity is applied to the output of the product unit.


\subsection{Hybrid summation-multiplication networks}
Both types of neurons can be combined in a hybrid summation-multiplication network.
Yet this poses the problem of how to distribute additive and multiplicative units over the network, i.e.\ how to determine whether a specific neuron should be an additive or multiplicative unit to obtain the best results.
A simple solution is to stack alternating layers of additive and product units, optionally with additional connections that skip over a product layer, so that each additive layer receives inputs from both the product layer and the additive layer beneath it.
The drawback of this approach is that the resulting uniform distribution of product units will hardly be ideal.

A more adaptive approach is to learn the function of each neural unit from provided training data.
However, since addition and multiplication are different operations, until now there was no obvious way to determine the best operation during training of the network using standard neural network optimization methods such as backpropagation.
An iterative algorithm to determine the optimal allocation could have the following structure:
For initialization randomly choose the operation each neuron performs.
Train the network by minimizing the error function and then evaluate its performance on a validation set.
Based on the performance determine a new allocation using a discrete optimization algorithm (such as particle swarm optimization or genetic algorithms).
Iterate the process until satisfactory performance is achieved.
The drawback of this method is its computational complexity; to evaluate one allocation of operations the whole network must be trained, which takes from minutes to hours for moderately sized problems.

Here we propose an alternative approach, where the distinction between additive and multiplicative neurons is not discrete but continuous and differentiable.
Hence the optimal distribution of additive and multiplicative units can be determined during standard gradient-based optimization.
Our approach is organized as follows:
First, we introduce non-integer iterates of the exponential function in the real and complex domains.
We then use these iterates to smoothly interpolate between addition \eqref{eq:additivenn} and multiplication \eqref{eq:multiplicativenn}.
Finally, we show how this interpolation can be integrated and implemented in neural networks.

\section{Iterates of the exponential function}

\subsection{Functional iteration}
Let $f: \setC \to \setC$ be an invertible function.
For $n \in \setZ$ we write $f^{(n)}$ for the n-times iterated application of $f$, 
\begin{equation}
f^{(n)}(z) = \underbrace{f \circ f \circ \cdots \circ f}_{n\text{ times}}(z)  \,. 
\label{eq:naturaliter}
\end{equation}
Further let $f^{(-n)} = (f^{-1})^{(n)}$ where $f^{-1}$ denotes the inverse of $f$.
We set $f^{(0)}(z) = z$ to be the identity function.
It can be easily verified that functional iteration with respect to the composition operator, i.e.
\begin{equation}
f^{(n)} \circ f^{(m)} = f^{(n+m)} 
\label{eq:itercomp}
\end{equation}
for $n,m \in \setZ$, forms an Abelian group.

Equation \eqref{eq:naturaliter} cannot be used to define functional iteration for non-integer $n$.
Thus, in order to calculate non-integer iterations of function, we have to find an alternative definition.
The sought generalization should also extend the additive property \eqref{eq:itercomp} of the composition operation to non-integer $n,m \in \setR$.

\subsection{Abel's functional equation}
\begin{figure}
\centering
\includegraphics[width=0.7\columnwidth,trim=0 15px 0 0]{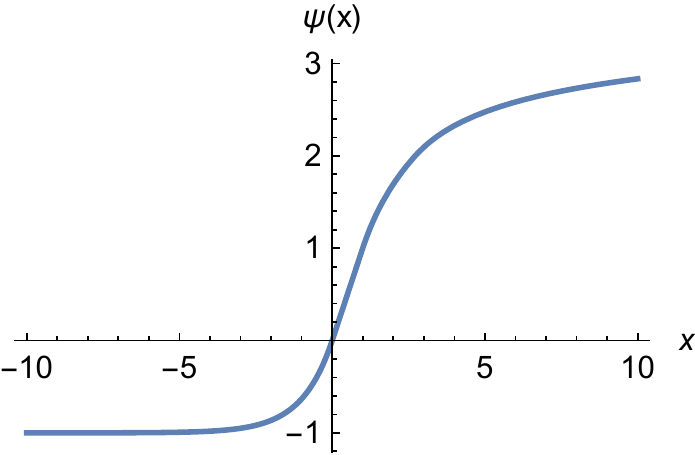}
\caption{
A continuously differentiable solution $\psi(x)$ to Abel's equation \eqref{eq:abel} for the exponential function in the real domain.
}
\label{fig:abelsol}
\end{figure}

Consider the following functional equation given by \cite{Abel1826},
\begin{equation}
\psi(f(x)) = \psi(x) + \beta
\label{eq:abel}
\end{equation}
with constant $\beta \in \setC$.
We are concerned with $f(x)=\exp(x)$.
A continuously differentiable solution for $\beta=1$ and $x \in \setR$ is given by
\begin{equation}
\psi(x) = \log^{(k)}(x) + k
\label{eq:abelsol}
\end{equation}
with $k \in \setN \text{ s.t. } 0 \leq \log^{(k)}(x) < 1$.
Note that for $x<0$ we have $k=-1$ and thus $\psi$ is well defined on whole $\setR$.
The function is shown in Fig.~\ref{fig:abelsol}.
Since $\psi: \setR \to (-1, \infty)$ is strictly increasing, the inverse $\psi^{-1}: (-1, \infty) \to \setR$ exists and is given by
\begin{equation}
\psi^{-1}(\psi) = \exp^{(k)}(\psi - k) \quad 
\label{eq:psiinv}
\end{equation}
with $k \in \setN \text{ s.t. } 0 \leq \psi-k < 1$.
For practical reasons we set $\psi^{-1}(\psi) = -\infty$ for $\psi \leq -1$. 
The derivative of $\psi$ is given by
\begin{subequations}
\label{eq:dpsi}
\begin{equation}
\psi'(x) = \prod_{j=0}^{k-1} \frac{1}{\log^{(j)}(x)}
\end{equation}
with $k \in \setN \text{ s.t. } 0 \leq \log^{(k)}(x) < 1$ and the derivative of its inverse is
\begin{equation}
\psi^{-1'}(\psi) = \prod_{j=0}^{k-1} \exp^{(j)}\!\left( \psi^{-1}(\psi - j)  \right) 
\end{equation}
with $k \in \setN \text{ s.t. } 0 \leq \psi -k < 1$.
\end{subequations}

\subsubsection{Non-integer iterates using Abel's equation}
By inspection of Abel's equation \eqref{eq:abel}, we see that the $n$th iterate of the exponential function can be written as
\begin{equation}
\exp^{(n)}(x) = \psi^{-1}\!\left( \psi(x) + n \right) \, .
\label{eq:abelexp}
\end{equation}
While this equation is equivalent to \eqref{eq:naturaliter} for integer $n$, we are now also free to choose $n \in \setR$ and thus \eqref{eq:abelexp} can be seen as a generalization of functional iteration to non-integer iterates.
It can easily be verified that the composition property \eqref{eq:itercomp} holds.
Hence we can understand the function $\varphi(x) = \exp^{(1/2)}(x)$ as the function that gives the exponential function when applied to itself. 
$\varphi$ is called the \emph{functional square root} of $\exp$ and we have $\varphi(\varphi(x)) = \exp(x)$ for all $x \in \setR$.
Likewise $\exp^{(1/N)}$ is the function that gives the exponential function when iterated $N$ times.

Since $n$ is a continuous parameter in definition \eqref{eq:abelexp} we can take the derivative of $\exp$ with respect to its argument as well as $n$.
They are given by
\begin{subequations}
\label{eq:dexp}
\begin{align}
\exp'^{(n)}(x) &= \frac{\partial \exp^{(n)}(x)}{\partial x} = \psi'^{-1}\!\left(\psi(x) + n \right) \psi'(x) \\
\exp^{(n')}(x) &= \frac{\partial \exp^{(n)}(x)}{\partial n} = \psi'^{-1}\!\left(\psi(x) + n \right) \,.
\end{align}
\end{subequations}
Thus \eqref{eq:abelsol} provides a method to interpolate between the exponential function, the identity function and the logarithm in a continuous and differentiable way.

\begin{figure}
\centering
\includegraphics[width=0.7\columnwidth,trim=0 15px 0 0]{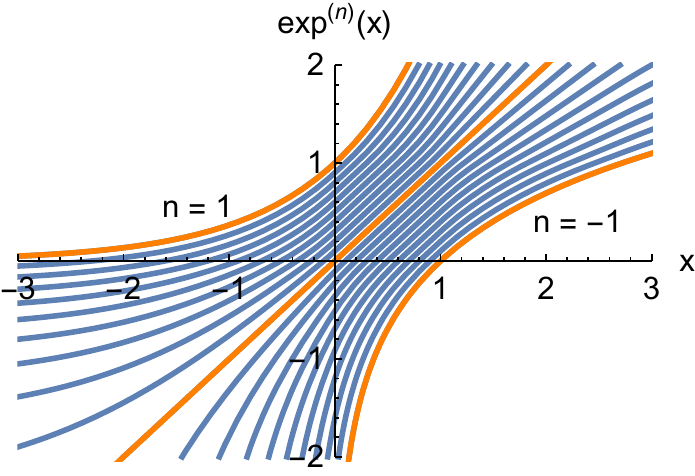}
\caption{
Iterates of the exponential function $\exp^{(n)}(x)$ for $n \in \{-1,-0.9,\dots,0,\dots,0.9,1\}$ obtained using the solution \eqref{eq:abelexp} of Abel's equation.
}
\label{fig:abelexp}
\end{figure}

\subsection{Schr\"oder's functional equation}
Motivated by the necessity to evaluate the logarithm for negative arguments, we derive a solution of Abel's equation for the \emph{complex} exponential function.
Applying the substitution 
\[ \psi(x) = \frac{\beta}{\log \gamma} \log \chi(x) \]
in Abel's equation \eqref{eq:abel} gives a functional equation first examined by \cite{Schroder1870},
\begin{equation}
\chi(f(z)) = \gamma \, \chi(z)
\label{eq:schroeder}
\end{equation}
with constant $\gamma \in \setC$.
As before we are interested in solutions of this equation for $f(x)=\exp(x)$; we have
\begin{equation}
\chi(\exp(z)) = \gamma \, \chi(z) \,
\label{eq:expschroeder}
\end{equation}
but now we are considering the complex $\exp: \setC \to \setC$.

The complex exponential function is not injective, since
\[ \exp(z + 2 \pi n i) = \exp(z) \quad \quad n \in \setZ \, .\]
Thus the imaginary part of the codomain of its inverse, i.e.\ the complex logarithm, must be restricted to an interval of size $2 \pi$.
Here we define $\log: \setC \to \{ z \in \setC: \beta \leq \Im z < \beta + 2\pi \}$ with $\beta \in \setR$.
For now let us consider the principal branch of the logarithm, that is $\beta = -\pi$.

To derive a solution, we examine the behavior of $\exp$ around one of its fixed points.
A fixed point of a function $f$ is a point $c$ with the property that $f(c) = c$.
The exponential function has an infinite number of fixed points.
Here we select the fixed point closest to the real axis in the upper complex half plane.
Since $\log$ is a contraction mapping, according to the Banach fixed-point theorem \cite{Khamsi2001} the fixed point of $\exp$ can be found by starting at an arbitrary point $z \in \setC$ with $\Im z \geq 0$ and repetitively applying the logarithm until convergence.
Numerically we find
\[ \exp(c) = c \approx 0.318132 + 1.33724 \, i \]
where $i = \sqrt{-1}$ is the imaginary unit.

Close enough to $c$ the exponential function behaves like an affine map.
To show this, let $z'=z-c$ and consider
\begin{align*}
\exp(c+z') -c &= \exp(c) \exp(z') -c  = c\, [ \exp(z') - 1] \\
			  &= c\, [ 1 + z' + O(|z'|^2) - 1 ] \\
			  &= c\, z' + O(|z'|^2) \,.
\end{align*}
Here we used the Taylor expansion of the exponential function, $\exp(z') = 1 + z' + O(z'^2)$.
Thus for any point $z$ in a circle of radius $r_0$ around $c$, we have
\begin{equation}
\exp(z) = cz + c - c^2 + O(r_0^2) \, .
\label{eq:expnearc}
\end{equation}
By substituting this approximation into \eqref{eq:expschroeder} it becomes apparent that a solution to Schr\"oder's equation around $c$ is given by 
\begin{equation}
\chi(z) = z - c \quad \text{for } |z- c| \leq r_0 
\label{eq:chinearc}
\end{equation}
where we have set $\gamma = c$.

We will now compute the continuation of the solution to points outside the circle around $c$.
From \eqref{eq:expschroeder} we obtain
\begin{equation}
\chi(z) = c \, \chi(\log(z)) \,.  
\label{eq:chiiter}
\end{equation}
If for a point $z \in \setC$ repeated application of the logarithm leads to a point inside the circle of radius $r_0$ around c, we can obtain the function value of $\chi(z)$ from \eqref{eq:chinearc} via iterated application of \eqref{eq:chiiter}.
In the next section it will be shown that this is indeed the case for nearly every $z \in \setC$.
Hence the solution to Schr\"oder's equation is given by
\begin{equation}
\chi(z) = c^k \, (\log^{(k)}(z) - c) 
\label{eq:chiexp}
\end{equation} 
with $k = \min_{k' \in \setN} k' \text{ s.t. } | \log^{(k')}(z) - c | \leq r_0$.
Solving for $z$ gives
\begin{equation}
\chi^{-1}(\chi) = \exp^{(k)} (c^{-k} \chi + c)
\label{eq:chi-1exp}
\end{equation}
with $k = \min_{k' \in \setN} k' \text{ s.t. } | c^{-k'} \chi | \leq r_0$.
Obviously we have $\chi^{-1}(\chi(z)) = z$ for all $z \in \setC$.
However $\chi(\chi^{-1}(\xi)) = \xi$ only holds if $\Im (c^{-k} \xi + c) \in [\beta, \beta+2\pi)$.

\begin{subequations}
\label{eq:dchi}
The derivative of $\chi$ is given by
\begin{equation}
\chi'(z) = \prod_{j=0}^{k-1} \frac{c}{\log^{(j)} z} 
\end{equation}
with $k = \min_{k' \in \setN} k' \text{ s.t. } | \log^{(k')}(z) - c | \leq r_0$ and we have
\begin{equation}
\chi'^{-1}(\chi) = \frac{1}{c} \prod_{j=1}^{k} \exp^{(j)}\!\left( \chi^{-1}\!\left(\frac{\chi}{c^{\,j}}\right) \right) 
\end{equation}
with $k = \min_{k' \in \setN} k' \text{ s.t. } | c^{-k'} \chi | \leq r_0$.
\end{subequations}

\begin{figure}
\begin{center}
\includegraphics[width=0.7\columnwidth,trim=0 10px 0 0]{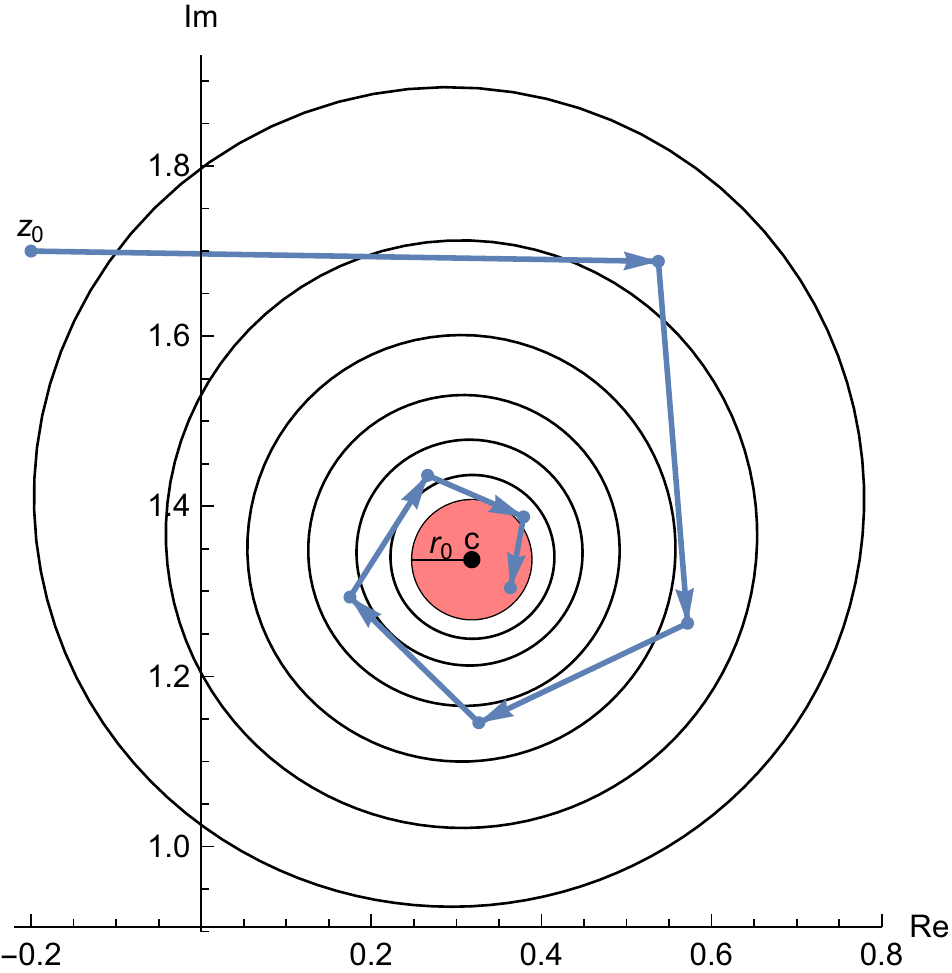}
\caption{
Calculation of $\chi(z)$.
Starting from point $z_0=z$ the series $z_{n+1} = \log z_n$ is evaluated until $|z_n - c| \leq r_0$ for some $n$. 
Inside this circle of radius $r_0$ the function value can then be evaluated using $\chi(z) = c^n (z_n -c)$.
The contours are generated by iterative application of $\exp$ to the circle of radius $r_0$ around $c$.
Near its fixed point the exponentiation behaves like a scaling by $|c| \approx 1.374$ and a rotation of $\Im c \approx 76.6^\circ$ around $c$.
}
\label{fig:chireccalc}
\end{center}
\end{figure}

\begin{figure}
\begin{center}
\includegraphics[width=0.9\columnwidth,trim=0 15px 0 0]{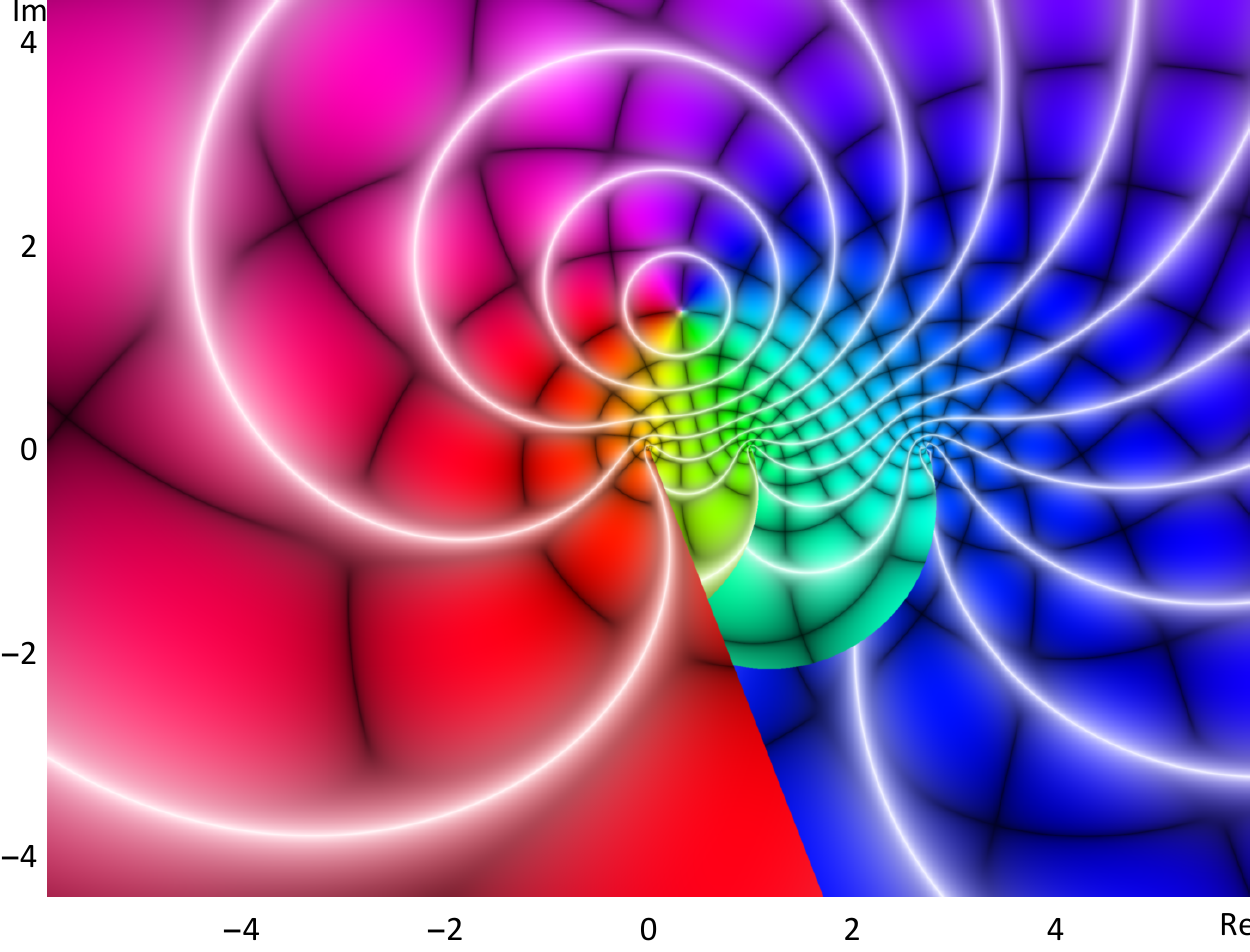}
\caption{
Domain coloring plot of $\chi(z)$.
Discontinuities arise at $0$, $1$, $\mathrm{e}, \mathrm{e}^\mathrm{e}, \dots$ and stretch into the negative complex half-plane.
They are caused by $\log$ being discontinuous at the polar angles $\beta$ and $\beta + 2\pi$.
}
\label{fig:chidomaincoloring}
\end{center}
\end{figure}

\subsubsection{The solution $\chi$ is defined on almost $\setC$}
\label{sec:chidomain}

The principal branch of the logarithm, i.e.\ restricting its imaginary part to the interval $[-\pi, \pi)$,  has the drawback that iterated application of $\log$ starting from a point on the lower complex half-plane will converge to the complex conjugate $\overline{c}$ instead of $c$.
Thus $\chi(z)$ would be undefined for $\Im z < 0$.

To avoid this problem, we use the branch defined by $\log: \setC \to \{ z \in \setC: \beta \leq \Im z < \beta + 2\pi \}$ with $-1 < \beta < 0$.
Using such a branch the series $z_n$, where 
\[ z_{n+1} = \log z_n \,,\]
converges to $c$, provided that there is no $n$ such that $z_n =0$.
Thus $\chi$ is defined on $\setC \setminus D$ where $D = \{0, \e, \e^\e, \e^{\e^\e}, \dots \}$.

\begin{proof}
If $\Im z_n \geq 0$ then $\arg z_n \in [0, \pi]$ and thus $\Im z_{n+1} \geq 0$.
Hence, if we have $\Im z_n \geq 0$ for $n \in \setN$, then $\Im z_{n'} \geq 0$ for all $n' > n$.
Now, consider the conformal map
\[ \xi(z) = \frac{z-c}{z-\overline{c}} \]
which maps the upper complex half-plane to the unit disk and define the series $\xi_{n+1} = \zeta(\xi_{n})$ with
\[ \zeta(t) = \xi\!\left( \log \xi^{-1}(t) \right)  \, . \]
We have $\zeta: D_1 \to D_1$, where $D_1 = \{ t \in \setC: |t| < 1 \}$ is the unit disk; furthermore $\zeta(0) = 0$.
Thus by Schwarz lemma $|\zeta(t)| < |t|$ for all $t \in D_1$ (since $\zeta(t) \neq \lambda t$ with $\lambda \in \setC$) and hence $\lim_{n\to\infty} \xi_n = 0$ \cite{Kneser1950}. 
This implies $\lim_{n\to\infty} z_n = c$.

On the other hand, if $\Im z_n < 0$ and $\Re z_n < 0$, then $\Im \log z_n > 0$ and $z_n$ converges as above.
Finally, if $\Im z_n < 0$ and $\Re z_n \geq 0$, then, using $-1 < \beta$, we have $\Re z_{n+1} \leq |\log z_n| \leq 1 + \log(\Re z_n) < \Re z_n$ and thus at some element $n'$ in the series we will have $\Re z_{n'} < 1$ which leads to $\Re z_{n'+1} < 0$.
\end{proof}

\subsubsection{Non-integer iterates using Schr\"oder's equation}
\begin{figure}
\centering
\includegraphics[width=0.49\columnwidth,trim=0 50px 0 0]{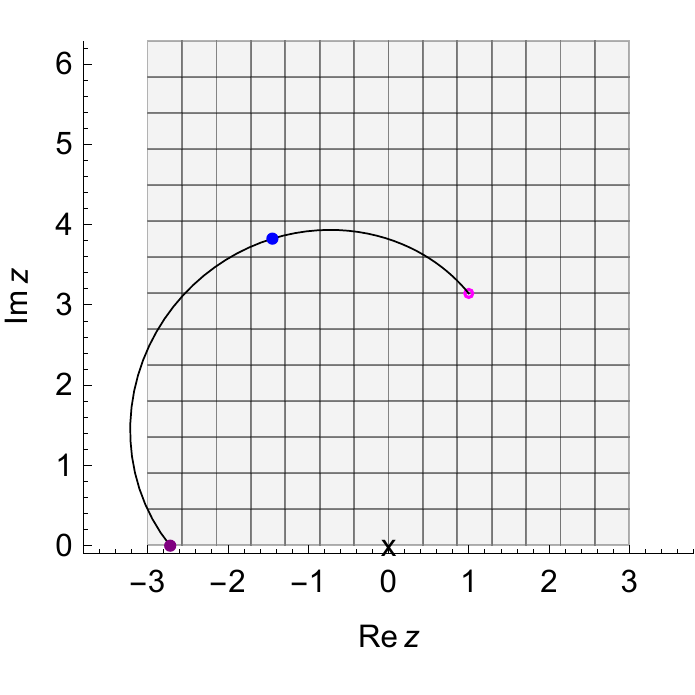} 
\includegraphics[width=0.49\columnwidth,trim=0 50px 0 0]{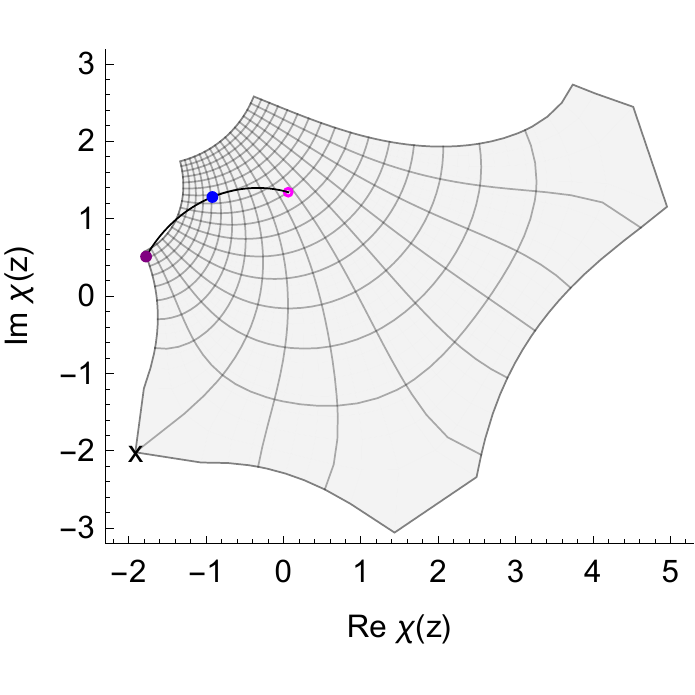}
\caption{
Structure of the function $\chi(z)$ and calculation of $\exp^{(n)}(1+\pi i)$ for $n \in [0,1]$ in $z$-space (left) and $\chi$-space (right).
The uniform grid with the cross at the origin is mapped using $\chi(z)$.
The points $1+\pi i$ and $\xi=\chi(1+\pi i)$ are shown as hollow magenta circles.
The black line shows $\chi^{-1}(c^n \, \xi)$ and $c^n \, \xi$ for $n \in [0,1]$.
The blue and purple points are placed at $n=1/2$ and $n=1$ respectively.
}
\label{fig:chimult}
\end{figure}

\begin{figure}
\centering
\includegraphics[width=1.0\columnwidth]{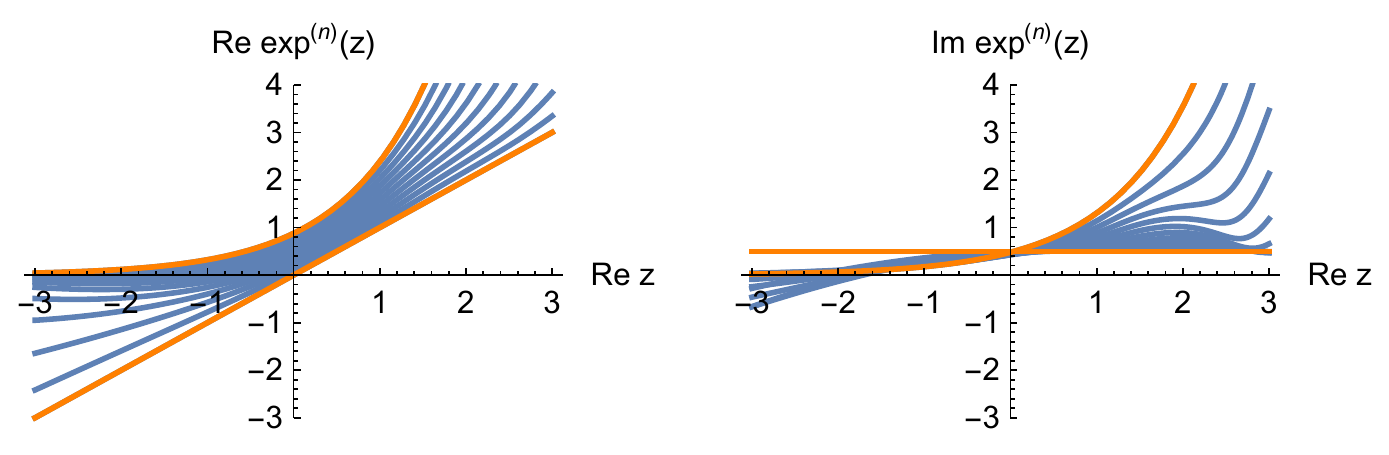} \\
\includegraphics[width=1.0\columnwidth,trim=0 40px 0 0]{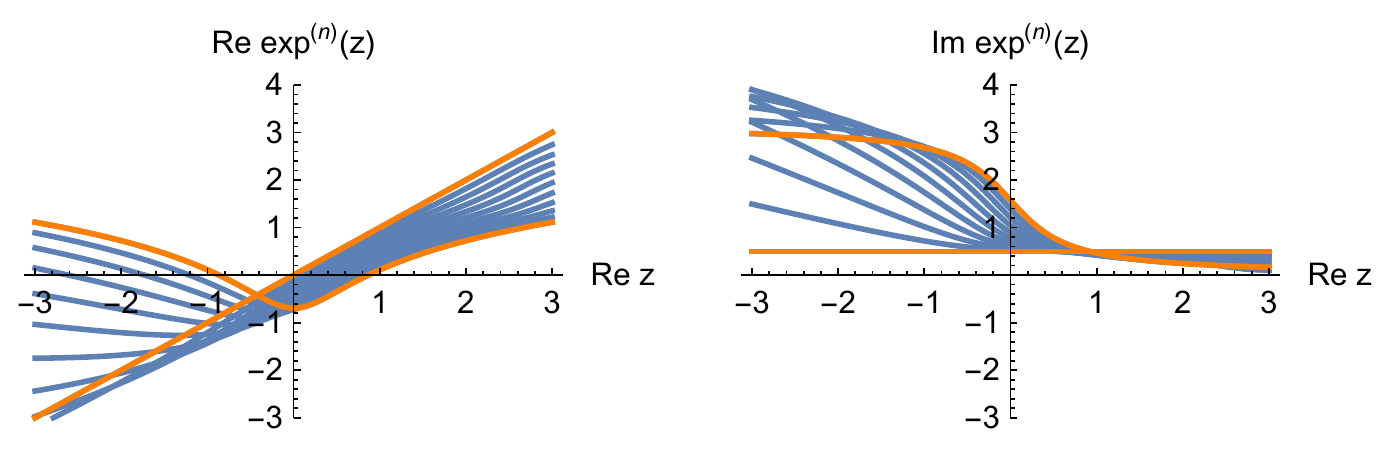}
\caption{
Iterates of the exponential function $\exp^{(n)}(x + 0.5i)$ for $n \in \{0,0.1,\dots,0.9,1\}$ (upper plots) and $n \in \{0,-0.1,\dots,-0.9,-1\}$ (lower plots) obtained using the solution \eqref{eq:abelexp} of Schr\"{o}der's equation.
Exp, log and the identity function are highlighted in orange.
}
\label{fig:schroederexp}
\end{figure}

Repetitive application of Schr\"oder's equation \eqref{eq:schroeder} on an iterated function \eqref{eq:naturaliter} leads to
\begin{equation}
\chi( f^{(n)}(z) ) = \gamma^n\, \chi(z) \,.
\end{equation}
Thus the $n$th iterate of the exponential function on the whole complex plane is given by
\begin{equation}
\exp^{(n)}(z) = \chi^{-1}(c^n \, \chi(z)) \, 
\label{eq:fracexp}
\end{equation}
where $\chi(z)$ and $\chi^{-1}(z)$ are given by \eqref{eq:chiexp} and \eqref{eq:chi-1exp} respectively.
Since $\chi$ is injective we can think of it as a mapping from the complex plane, called $z$-plane, to another complex plane, called $\chi$-plane.
By \eqref{eq:fracexp} the operation of calculating the exponential  of a number $y$ in the $z$-plane corresponds to complex multiplication by factor $c$ of $\chi(y)$ in the $\chi$-plane. 
This is illustrated in Fig.~\ref{fig:chimult}.
Samples from $\exp^{(n)}$ are shown in Fig.~\ref{fig:schroederexp}.

\begin{figure}
\centering
\includegraphics[width=0.75\columnwidth,trim=0 20px 0 0]{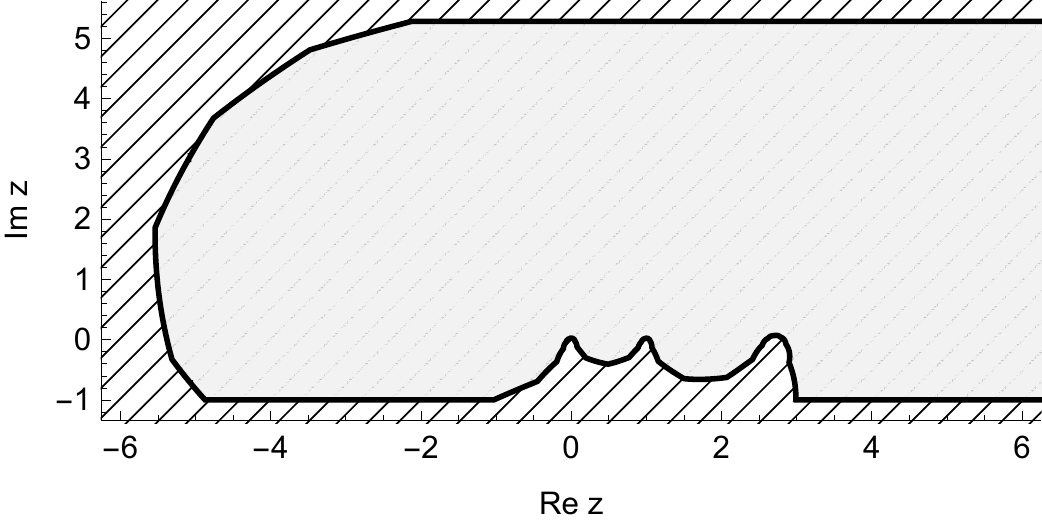}
\caption{
Function composition holds in the gray-shaded area for non-integer iteration numbers, i.e.\! $\exp^{(n)} \circ \exp^{(m)} (z) = \exp^{(n+m)}(z)$ for $n \in \setR$ and $m \in [-1,1]$.
We defined $\log$ such that $\Im \log z \in [-1, -1+2\pi]$.
}
\label{fig:comparea}
\end{figure}

While the definition for $\exp^{(n)}$ given by \eqref{eq:fracexp} can be evaluated on the whole complex plane $\setC$, it only has meaning as a non-integer iterate of $\exp$, if composition $ \exp^{(n)}\!\left[ \exp^{m}(z) \right]  = \exp^{(n+m)}(z) $ holds.
Since this requires that $\chi(\chi^{-1}(\xi)) = \xi$, let us define the sets $\mathcal{E}' = \{\xi \in \setC: \chi[\chi^{-1}(c^m \, \xi)] = c^m \, \xi \,\, \forall m \in [-1, 1] \}$ and $\mathcal{E} = \chi^{-1}(\mathcal{E}')$.
Then, for $z \in \mathcal{E}$, $n \in \setR$ and $m \in [-1,1]$, the composition of the iterated exponential function is given by
\begin{align*}
\exp^{(n)}[\exp^{(m)}(z)] &= \chi^{-1}\!\left[ c^n \, \chi\!\left( \chi^{-1}(c^m \, \chi(z) ) \right) \right]  \\
 &= \chi^{-1}\!\left[c^{n+m} \, \chi(z) \right]  = \exp^{(n+m)}(z) \,
\end{align*}
and the composition property is satisfied.
The subset $\mathcal{E}$ of the complex plane where composition of $\exp^{(n)}$ for non-integer $n$ is shown in figure \ref{fig:comparea}.

The derivatives of $\exp^{(n)}$ defined using Schr\"{o}der's equation are given by
\begin{subequations}
\label{eq:dexp_schroeder}
\begin{align}
\exp'^{(n)}(z) &= c^n \, \chi'^{-1}[ c^n \chi(z) ] \, \chi'(z) \\
\exp^{(n')}(z) &= c^n \, \chi'^{-1}[c^n \chi(z)] \, \chi(z) \log(c) \,.
\end{align}
\end{subequations}

Hence we defined the continuously differentiable function $\exp^{(n)}: \setC \setminus D \to \setC$ on almost the whole complex plane and showed that it has the meaning of a non-integer iterate of $\exp$ on the subset $\mathcal{E}$.

\section{Interpolation between addition and multiplication}
Using fundamental properties of the exponential function we can write every multiplication of two numbers $x, y \in \setR$ as
\[ x y = \exp(\log x + \log y) = \exp(\exp^{-1} x + \exp^{-1} y)\,. \]
We define the operator $\oplus_n$ for $x,y \in \setR$ and $n \in \setR$ as
\begin{equation}
x \oplus_n y = \exp^{(n)}\!\left( \exp^{(-n)}(x) + \exp^{(-n)}(y) \right) \, .
\label{eq:addiplication}
\end{equation}
Note that we have $x \oplus_0 y = x + y$ and $x \oplus_1 y = x y$.
Thus for $0 < n < 1$ the above operator continuously interpolates between the elementary operations of addition and multiplication.
We will refer to $\oplus_n$ as the ``addiplication operator''.
Analogous to the n-ary sum and product we will employ the following notation for the n-ary addiplication operator,
\begin{equation}
\bigoplus_{\substack{j=k\\n}}^K x_j = x_k \oplus_n x_{k+1} \oplus_n \dots \oplus_n x_K \,.
\end{equation}

The derivative of the addiplication operator w.r.t.\ its operands and the interpolation parameter $n$ are calculated using the chain rule.
\begin{subequations}
Using the shorthand
\begin{align}
E =& \exp^{(-n)}(x) + \exp^{(-n)}(y) 
\end{align}
we have
\begin{align}
\frac{\partial (x \oplus_n y)}{\partial x} =& \exp'^{(n)}\!\left( E \right) \exp'^{(-n)}(x) \\
\frac{\partial (x \oplus_n y)}{\partial y} =& \exp'^{(n)}\!\left( E \right) \exp'^{(-n)}(y) \\
\frac{\partial (x \oplus_n y)}{\partial n} =& \exp^{(n')}\!\left( E \right) +  \\
 										    & \exp'^{(n)}\!\left( E \right) \cdot \left[ \exp^{(-n')}(x) + \exp^{(-n')}(y) \right] \nonumber \,.
\end{align}
\label{eq:daddiplication}
\end{subequations}

\begin{figure}
\centering
\includegraphics[width=0.9\columnwidth,trim=0 10px 0 0]{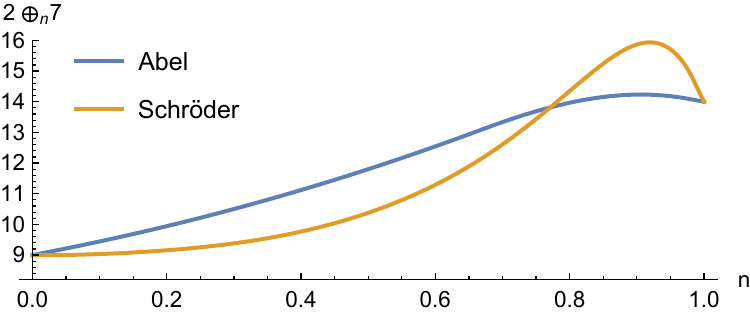}
\caption{
The interpolation between addition and multiplication using \eqref{eq:addiplication} with $x=2$ and $y=7$.
The iterates of the exponential function are either calculated using Abel's equation (blue) or Schr\"{o}der's equation (orange).
In both cases the interpolated values exceed the range between $2+7=9$ and $2\cdot 7 =14$ and therefore a local maximum exists.
}
\label{fig:addiplicaton}
\end{figure}

For positive arguments $x,y>0$ we can use the iterates of $\exp$ based either on the solution of Abel's equation \eqref{eq:abelexp} or Schr\"{o}der's equation \eqref{eq:fracexp}.
However, if we also want to deal with negative arguments, we must use iterates of $\exp$ based on Schr\"{o}der's equation \eqref{eq:fracexp}, since the real logarithm is only defined for positive arguments.
From the exemplary addiplication shown in Fig.~\ref{fig:addiplicaton} we can see that the interpolations produced by these two methods are not monotonic functions w.r.t.\ the interpolation parameter $n$.
In both cases local maxima exist; however interpolation based on Schr\"oder's equation has higher extrema in this case and also in general (personal experiments).
It is well known, that the existence of local extrema can pose a problem for gradient-based optimizers.

\section{Neurons that can add or multiply}

We propose two methods to construct neural nets that have units the operation of which can be adjusted using a continuous parameter.

\begin{figure}
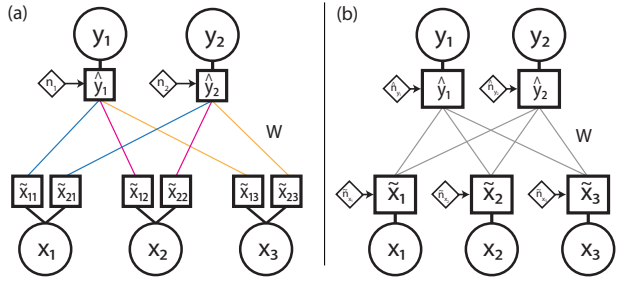

\centering
\includegraphics[height=3.5cm,trim=0 10px 0 0]{net_addiplication.ai}
\hfill
\rulesep
\includegraphics[height=3.5cm,trim=0 10px 0 0]{net_logexp.ai}
\caption{
Two proposed neural network architectures that can implement addiplication.
(a) Neuron $y_i$ calculates its value according to \eqref{eq:nn_addiplication}.
We have $y_i = \sigma(\hat{y}_i)$ and the subunits compute $\tilde{x}_{ij} = \exp^{(-n_i)}(x_j)$ and $\smash{\hat{y}_i = \exp^{(n_i)}\!\left( \sum_j W_{ij} \tilde{x}_{ij} \right)}$.
The weights between the subunits are shared.
(b) Neuron $y_i$ calculates its value according to \eqref{eq:nn_logexp}.
We have $y_i = \sigma(\hat{y}_i)$ and the subunits compute $\tilde{x}_{j} = \exp^{(-\tilde{n}_j)}(x_j)$ and $\smash{\hat{y}_i = \exp^{(\hat{n}_i)}\!\left( \sum_j W_{ij} \tilde{x}_{j} \right)}$.
}
\label{fig:adpnet}
\end{figure}

The straightforward approach is to use neurons that use addiplication instead of summation, i.e.\ the value of neuron $y_i$ is given by
\begin{align}
y_i =\,& \sigma\!\left( \bigoplus_{\substack{j\\n_i}} W_{ij} x_j \right) \nonumber \\
    =\,& \sigma\!\left[ \exp^{(n_i)}\!\left( \sum_j W_{ij} \exp^{(-n_i)}(x_j) \right) \right] \,. 
\label{eq:nn_addiplication}
\end{align}
For $n_i=0$ the neuron behaves like an additive neuron and for $n_i=1$ it computes the product of its inputs.
Because we sum over $\exp^{(-n_i)}(x_j)$ which has dependence on the parameter $n_i$ of neuron $y_i$, this calculation corresponds to a network in which each neuron in layer $\vec{x}$ has separate outputs for each neuron in the following layer $\vec{y}$; see Fig.~\ref{fig:adpnet}a.
Compared to conventional neural nets this architecture has only one additional real-valued parameter per neuron ($n_i$) but also poses a significant increase in computational complexity due to the necessity of separate outputs.
Since $\exp^{(-n_i)}(x_j)$ is complex it might be sensible (but is not required) to allow a complex weight matrix $W_{ij}$.

The computational complexity of separate output units can be avoided by calculating the value of a neuron according to
\begin{equation}
y_i = \sigma\!\left[ \exp^{(\hat{n}_{y_i})}\!\left( \sum_j W_{ij} \exp^{(\tilde{n}_{x_j})}(x_j) \right) \right] \,. 
\label{eq:nn_logexp}
\end{equation}
This corresponds to the architecture shown in Fig.~\ref{fig:adpnet}b.
The interpolation parameter $n_i$ has been split into a pre-transfer-function part $\hat{n}_{y_i}$ and a post-transfer-function part $\tilde{n}_{x_j}$.
Since $\hat{n}_{y_i}$ and $\tilde{n}_{x_j}$ are not tied together, the network is free to implement arbitrary combinations of iterates of the exponential function.
Addiplication occurs as the special case $\hat{n}_{y_i} = -\tilde{n}_{x_j}$.
Compared to conventional neural nets each neuron has two additional parameters, namely $\hat{n}_{y_i}$ and $\tilde{n}_{x_j}$; however the asymptotic computational complexity of the network is unchanged.
In fact, this architecture corresponds to a conventional, additive neural net, as defined by \eqref{eq:additivenn}, with a neuron-dependent, parameterizable transfer function.
For neuron $z_i$ the transfer function given by
\begin{equation}
\sigma_{z_i}(t) =  \exp^{(\tilde{n}_{z_i})}\!\left[ \sigma\!\left( \exp^{(\hat{n}_{z_i})}(t) \right)  \right] \,.
\end{equation}
Consequently, implementation in existing neural network frameworks is possible by replacing the standard sigmoidal transfer function with this function and optionally using a complex weight matrix.

\section{Applications}
\subsection{Variable pattern shift}
\begin{figure}
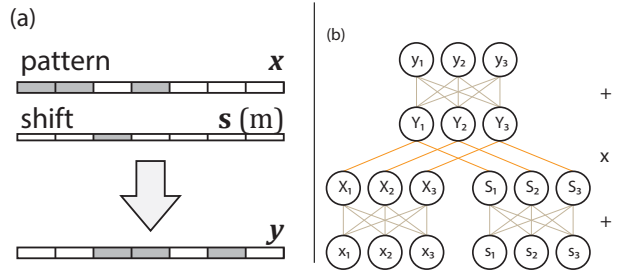

\centering
\includegraphics[width=0.47\columnwidth]{patternshift.ai}
\hfill
\rulesep
\includegraphics[width=0.47\columnwidth,trim=0 10px 0 0]{patternshift_net.ai}
\caption{
(a)
Variable pattern shift problem.
Given a random binary pattern $\vec{x} \in \setR^N$ and an integer $m \in \{0, 1, \dots, N-1\}$ presented in one-hot encoding, the learner should output the pattern $\vec{x}$ circularly shifted to the right by $m$ grid cells.
(b) 
A neural net with two hidden layers that can solve this problem by employing the Fourier shift theorem.
The first hidden layer is additive, the second is multiplicative; the output layer is additive.
All neurons use linear transfer functions.
The first hidden layer computes the DFT of the input pattern $\vec{x}$ and shift mount $\vec{s}$.
The second hidden layer applies the Fourier shift theorem and the output layer computes the inverse DFT of the shifted pattern. 
}
\label{fig:shiftproblem}
\end{figure}

Consider the dynamic pattern shift task shown in Fig.~\ref{fig:shiftproblem}a:
The input consists of a binary vector $\vec{x}$ of $N$ elements and an integer $m \in \{0, 1, \dots, N-1\}$.
The desired output $\vec{y}$ is $\vec{x}$ circularly shifted by $m$ elements to the right,
\[ y_n = x_{n-m} \,, \]
where $\vec{x}$ is indexed modulo $N$, i.e. $x_{n-m}$ rolls over to the right if $n-m \leq 0$.

A method to architecturally efficiently implement this task in a neural architecture is based on the shift theorem of the discrete Fourier transform (DFT) \cite{Brigham1988}.
Let $\mathcal{F}(\{x_n\})_k$ denote the $k$th element of the DFT of $\vec{x}=\{x_0, \dots, x_{N-1}\}$.
By definition we have 
\[ \mathcal{F}(\{x_n\})_k = \sum_{n=0}^{N-1} x_n \, \e^{- 2 \pi i \frac{kn}{N}}  \]
and its inverse is given by
\[ \mathcal{F}^{-1}(\{X_k\})_n = \frac{1}{N} \sum_{k=0}^{N-1} X_k \, \e^{2\pi i \frac{k n}{N}} \, .\]
The shift theorem states that a shift by $m$ elements in the time domain is equivalent to a multiplication by factor $\e^{-2 \pi i k m / N}$ in the frequency domain,
\[ \mathcal{F}(\{x_{n-m}\})_k = \mathcal{F}(\{x_n\})_k \, \e^{-2 \pi i \frac{k m}{N}} \,. \]
Hence the shifted pattern can be calculated using
\[ \vec{y} = \mathcal{F}^{-1}(\{ \mathcal{F}(\{x_n\})_k \, \e^{-2 \pi i \frac{k m}{N}} \}) \,.\]
Using the above definitions its $v$th component is given by
\[ y_v =  \frac{1}{N} \sum_{k=0}^{N-1} \e^{2\pi i \frac{k v}{N}} \left[ \e^{-2 \pi i \frac{k m}{N}} \sum_{n=0}^{N-1} x_n \, \e^{- 2 \pi i \frac{kn}{N}} \right]  \,. \]
If we encode the shift amount $m$ as a one-hot vector $\vec{s}$ of length $N$, i.e.\ $s_j=1$ if $j=m$ else $s_j=0$, we can further rewrite this as
\begin{subequations}
\begin{align}
y_v &=  \frac{1}{N} \sum_{k=0}^{N-1} \e^{2\pi i \frac{k v}{N}} \, S_k \, X_k 
\end{align}
with
\begin{align}
S_k = \sum_{m=0}^{N-1} s_m \,\e^{-2 \pi i \frac{k m}{N}} \,, \quad 
X_k = \sum_{n=0}^{N-1} x_n \, \e^{- 2 \pi i \frac{kn}{N}} \,.
\end{align}
\label{eq:shiftnet_sol}
\end{subequations}
This corresponds to a neural network with two hidden layers (one additive, one multiplicative) and an additive output layer as shown in Fig.~\ref{fig:shiftproblem}b.
The optimal weights of this network are given by the corresponding coefficients from \eqref{eq:shiftnet_sol}.

From this example we see that having the ability to automatically determine the function of each neuron is crucial to learn neural nets that are able to solve complex problems.

\section{Conclusion and future work}
We proposed one method to continuously and differentiably interpolate between addition and multiplication and showed how it can be integrated into neural networks by replacing the standard sigmoidal transfer function with a parameterizable transfer function.
In this paper we presented the mathematical formulation of these concepts and showed how to integrate them into neural networks.

We will perform simulations to see how these neural nets behave given real-world problems and how to train them most efficiently.

While working on this theory, we already have two possible improvements in mind: 
\begin{enumerate}
\item
Our interpolation technique is based on non-integer iterates of the exponential function calculated using Abel's and Schr\"oder's functional equations.
We chose this method for our first explorations because it has the mathematically sound property that the calculated iterates form an Abelian group under functional composition.
However it results in a non-monotonic interpolation between addition and multiplication which may lead to a challenging optimization landscape during training.
Therefore we will try to find more ad-hoc interpolations with a monotonic transition between addition and multiplication.

\item
Our method introduces one or two additional real-valued parameters per neuron for the transfer function.
Using a suitable fixed transfer function might allow to absorb these parameters back into the bias.
\end{enumerate}

While the specific implementation details proposed in this work may have their drawbacks, we believe that neurons which can implement operations beyond addition are a key to new areas of application for neural computation.

\section*{Acknowledgments}
Part of this work has been supported in part by the TAC-MAN project, EC Grant agreement no. 610967, within the FP7 framework programme.

\bibliography{addiplication}
\bibliographystyle{icml2013}
\end{document}